\documentclass[10pt,twocolumn,letterpaper]{article}

\usepackage{iccv}
\usepackage{times}
\usepackage{epsfig}
\usepackage{graphicx}
\usepackage{amsmath}
\usepackage{amssymb}
\usepackage{wrapfig}
\usepackage{subcaption}
\usepackage{color}
\usepackage{array}
\usepackage{makecell}
\graphicspath{{f/}}

\usepackage[breaklinks=true,bookmarks=false]{hyperref}

\iccvfinalcopy 

\def\iccvPaperID{2} 

\newcommand{\myparagraph}[1]{\smallskip\noindent\textbf{#1}}

\ificcvfinal\fi
\begin{document}
	
	\title{\LaTeX\ Author Guidelines for ICCV Proceedings}
	
	\title{3D Pose Regression using Convolutional Neural Networks}
	
	\author{Siddharth Mahendran \\
		{\tt\small siddharthm@jhu.edu}
		\and
		Haider Ali \\
		{\tt\small hali@jhu.edu}
		\and
		Ren{\'e} Vidal \\
		{\tt\small rvidal@cis.jhu.edu}
		\and
		Center for Imaging Science, Johns Hopkins University
		\vspace{-2mm}
	}
	
	\maketitle

	\begin{abstract}
		3D pose estimation is a key component of many important computer vision tasks such as autonomous navigation and 3D scene understanding. Most state-of-the-art approaches to 3D pose estimation solve this problem as a pose-classification problem in which the pose space is discretized into bins and a CNN classifier is used to predict a pose bin. We argue that the 3D pose space is continuous and propose to solve the pose estimation problem in a CNN regression framework with a suitable representation, data augmentation and loss function that captures the geometry of the pose space. Experiments on PASCAL3D+ show that the proposed 3D pose regression approach achieves competitive performance compared to the state-of-the-art.
	\end{abstract}
	
	\section{Introduction}
	\label{sec:intro}
	A 2D image is a snapshot of the 3D world and retrieving 3D information from images is an old and fundamental challenge in computer vision. One way to describe the underlying 3D scene is to report the 3D pose of all the objects present in the scene. This task is known as 3D pose estimation and it is a key component of vision problems such as scene understanding and 3D reconstruction. It also plays a vital role in modern vision challenges such as autonomous navigation, where the ability to quickly and reliably recover the 3D pose of other automobiles, pedestrians and objects relative to the camera is very important. 
	
	The term 3D pose refers to the transformation between the object and the camera and is often captured using 6 parameters: azimuth $az$, elevation $el$, camera-tilt $ct$, distance to the camera $d$ and image translation $(px, py)$. In this work however, we are not interested in the full 6-dof pose, but in the rotation transformation $R$ between the object and the camera, which is captured by the first three parameters, i.e. $R(az, el, ct)$. Note that we also make a distinction between the tasks of 3D pose estimation and 2D detection. We assume that we have the output of a 2D detection system or an oracle that gives us a bounding box around the object in an image. We then process the image patch inside the bounding box to predict the rotation $R$. We do this by using a deep convolutional neural network (CNN) that regresses the 3D pose given this 2D image patch. 
	
	\myparagraph{Related Work} There is a rich literature on 3D pose estimation from a single image, from the earlier work of \cite{Schneiderman-Kanade:CVPR00} to the more recent work of \cite{Pepik:CVPR12, Hejrati:CVPR14}. Due to space constraints, we concentrate our review on CNN-based methods, which can be grouped into two categories. Methods in the first category, such as \cite{Wu:ECCV16} and \cite{Pavlakos:ICRA17}, predict 2D keypoints from an image and then use 3D object models to predict the 3D pose given these keypoints. Methods in the second category, such as Viewpoints and Keypoints (V\&K)\cite{Tulsiani:CVPR15} and Render-for-CNN \cite{Su:ICCV15}, which are closer to what we do, predict 3D pose directly given an image. Both of these methods discretize the pose space into bins and solve a pose classification problem. They have a similar network architecture, which is shared across object categories up to the second-last layer and a separate output layer for every category. While V\&K \cite{Tulsiani:CVPR15} uses a standard cross-entropy loss for classification, Render-for-CNN \cite{Su:ICCV15} uses a weighted cross-entropy loss that respects the circular symmetry of angles. While V\&K \cite{Tulsiani:CVPR15} uses jittered bounding boxes with sufficient overlap to augmented annotated training data, Render-for-CNN \cite{Su:ICCV15} uses rendered images with a well-sampled distribution over pose space, random crops, and backgrounds. Another method in the second category is \cite{Elhoseiny:ICML16}, which studies multiview CNN models for joint object categorization and pose estimation, and their models also solve for pose labels. 
	
	\myparagraph{Contributions} In this work, we argue that since \textit{the 3D pose space is continuous, the pose estimation problem can be solved in a regression framework} rather than breaking up the pose space into discrete bins. The challenge is that the 3D pose space is non-Euclidean, hence CNN algorithms need to be modified to account for the nonlinear structure of the output space. Our key contribution is to develop a CNN regression framework for solving the 3D pose estimation problem in the continuous domain by designing a suitable representation, data augmentation and loss function that respect the non-linear structure of the 3D pose space.
	
	\begin{table*}[ht]
		\centering
		\begin{tabular}{|c|c|c|c|}
			\hline
			& V\&K \cite{Tulsiani:CVPR15} & Render-for-CNN \cite{Su:ICCV15} & Ours \\
			\hline
			Problem formulation & Classification & Fine-grained classification & Regression \\
			\hline
			Representation & Discretized angles (21 bins) & Discretized angles (360 bins) & Axis-angle / Quaternion \\
			\hline
			Loss function & Cross-entropy & Weighted cross-entropy & Geodesic loss \\
			\hline
			Data augmentation & 2D jittering & Rendered images &3D pose jittering + rendered images \\
			\hline
			Network architecture & VGG-Net (FC7) & AlexNet (FC7) & VGG-M (FC6) \\
			\hline
		\end{tabular}
		\caption{A comparison of the state-of-the-art methods and our proposed framework}
		\label{table:contributions}
	\end{table*}

	 More specifically, we use a modified VGG network architecture that consists of a feature network that is shared between all object categories and a pose network that is specific to each category. The pose network models 3D pose using an appropriate representation, non-linearity and loss function. We study two representations in particular, axis-angle and quaternions, and model their constraints using non-linearities in the output layer. Our loss function is a geodesic distance on the space of rotation matrices. We also propose a data augmentation technique that is more suitable for regression compared to jittering. We also present experiments on the Pascal3D+ dataset together with an ablation analysis of our various design choices, which show competitive performance with respect to state-of-the-art methods. We present a comparison of our proposed framework with current state-of-the-art methods in Table~\ref{table:contributions}.
	
	To the best of our knowledge, our work is the first one that does 3D object pose regression using CNNs with axis-angle/quaternion representations and geodesic loss functions, and shows good performance on a challenging dataset like Pascal3D+ \cite{Xiang:WACV14}. We also note that 3D pose regression is commonly used in human pose estimation, to regress the joint locations of the human skeleton. Quaternions have also been used to represent 3D pose for camera localization in \cite{Kendall:ICCV15, Kendall:CVPR17, Kendall:ICRA16}, but these works ignore the unit-norm constraint for computational ease and use a mean-squared or reprojection loss, whereas we incorporate the constraint into the network and use a geodesic loss.
	
	\section{3D Pose Regression using CNNs}
	\label{sec:model}
	In this section, we describe our regression framework in detail. We first describe the two 
	representations of 3D rotation matrices we use: axis-angle and quaternions, and the 
	corresponding non-linear activations and loss functions. We then describe our network architecture. Finally, we present our proposed data augmentation strategy.
	
	\subsection{Representing 3D Rotations}
	\label{sec:representation}
	Any rotation matrix $R$ lies in the set of special orthogonal matrices $SO(3) \doteq \{ R: R 
	\in \mathbb{R}^{3 \times 3}, R^T R = I_3, \det(R) = 1 \}$. We can then define a geodesic 
	distance between two rotation matrices, $R_1$ and $R_2$ as shown in Eqn.~\eqref{eq:geodesicR}, where $\log$ is the matrix logarithm and $\|\cdot\|_F$ is the Frobenius norm. This is also the loss function we use in our networks, simplified depending on the representation. 
	\begin{equation}
	d(R_1, R_2) = \frac{\|\log(R_1 R_2^T) \|_F}{\sqrt{2}} .
	\label{eq:geodesicR}
	\end{equation}
	
	\myparagraph{Axis-angle} A rotation matrix $R$ captures the rotation of 3D points by an angle 
	$\theta$ about an axis $v, \|v\|_2=1$. This can be expressed as $R = \exp(\theta [v]_{\times})$, where $\exp$ is the matrix exponential and $[v]_\times$ is the skew-symmetric operator of vector $v$, i.e, $[v]_\times = \begin{pmatrix}
	0 & -v_3 & v_2 \\ v_3 & 0 & -v_1 \\ -v_2 & v_1 & 0 
	\end{pmatrix}$ for $v = [v_1, v_2, v_3]^T$. So, every rotation matrix $R$ has a corresponding 
	axis-angle vector $y = \theta v$ and vice-versa. We also restrict $\theta \in [0, \pi)$ and 
	define $R=I_3 \Leftrightarrow y=0_3$, which ensures a unique mapping between rotation matrix $R$ and it's representation $y$. The matrix exponential can be simplified to $R = I_3 + 
	\sin \theta [v]_\times + (1 - \cos \theta) [v]_\times^2$ using the Rodrigues' rotation formula. 
	In the same way, Eqn.~\eqref{eq:geodesicR} can be simplified to get: 
	\begin{equation}
	d_A(R_1, R_2) = \cos^{-1} \left[ \frac{tr(R_1^T R_2) - 1}{2} \right] .
	\label{eq:distR}
	\end{equation}
	Note that $\| \log \big (\exp (\theta_1 [v_1]_\times) \exp (\theta_2 [v_2]_\times)^T \big ) \|_F / 
	\sqrt{2}$ looks very similar to $\|\theta_1 v_1 - \theta_2 v_2\|_2$, but it is not the same 
	because $\exp(\theta_1 [v_1]_\times) \exp(\theta_2 [v_2]_\times)^T \neq \exp(\theta_1 [v_1]_\times - \theta_2 [v_2]_\times) $ in general. The equality holds only when the matrices $[v_1]_\times$ and $[v_2]_\times$ commute i.e. $v_1 = \pm v_2$.
	
	\myparagraph{Quaternion} Another popular representation for 3D rotation matrices are 
	quaternions. Given an axis-angle vector $y = \theta v$, the corresponding quaternion $q = (c, s)$ is given by $(\cos \frac{\theta}{2}, \sin \frac{\theta}{2} v)^T$. By construction, quaternions are unit-norm, $\|q\|_2=1$. Using quaternion algebra, we have $(c_1, s_1).(c_2, s_2) = (c_1c_2- \langle s_1, s_2 \rangle, c_1 s_2+c_2 s_1+s_1 \times s_2)$ and $(c, s)^{-1} = (c, -s)$ for unit norm $q=(c, s)$. Now, expressing Eqn.~\eqref{eq:geodesicR} in terms of quaternions $q_1$ and $q_2$, we have:
	\begin{equation}
	d(q_1, q_2) = 2 \cos^{-1}(|c|) \hspace{1em} \text{where} \hspace{1em} (c, s) = q_1^{-1} \cdot q_2
	\end{equation}
	which we simplify to get:
	\begin{equation}
	d_Q(q_1, q_2) = 2 \cos^{-1}(| \langle q_1, q_2 \rangle|) .
	\label{eq:distQ}
	\end{equation}
	
	\subsection{Network Architecture}
	\label{sec:architecture}

\begin{figure}[h]
		\centering
		\includegraphics[width=\linewidth]{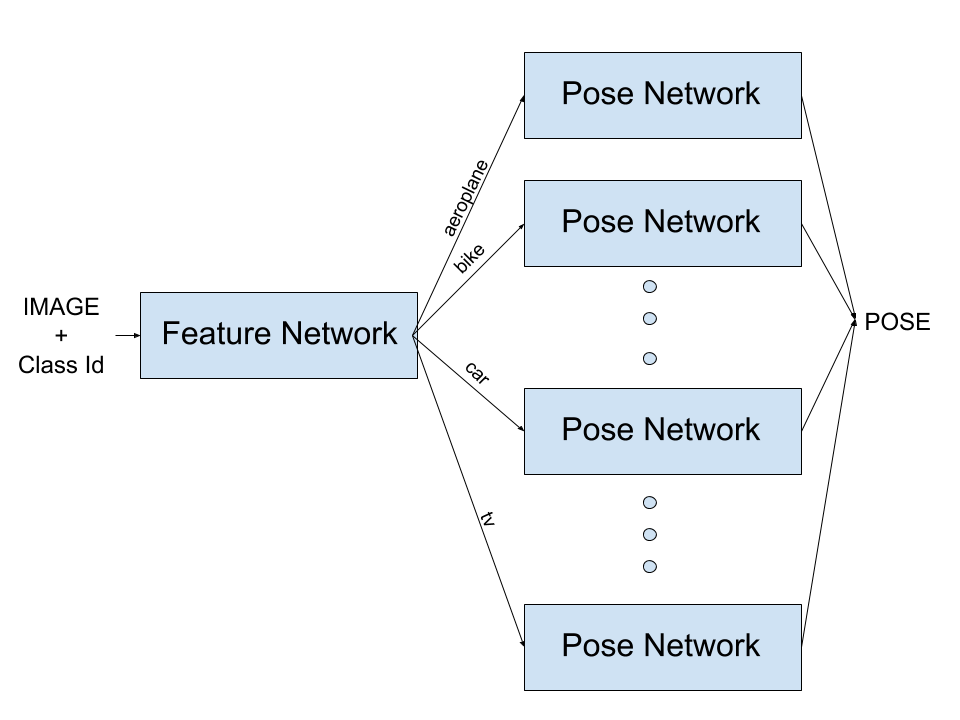}
		\caption{Overall network architecture where the Feature Network is shared across object 
			categories while each category has its own Pose Network.}
		\label{fig:overall_network}
	\end{figure}

\begin{figure}[h]
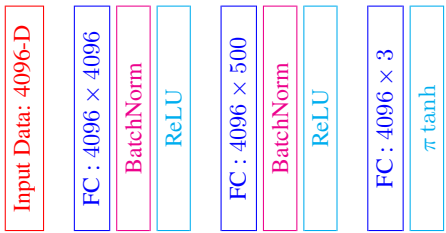

		\small
		\centering
		{\color{red}\rotatebox{90}{\framebox[3cm][c]{Input Data: 4096-D}}}
		\quad
		{\color{blue}\rotatebox{90}{\framebox[3cm][c]{FC : $4096 \times 4096$}}}
		{\color{magenta}\rotatebox{90}{\framebox[3cm][c]{BatchNorm}}}
		{\color{cyan}\rotatebox{90}{\framebox[3cm][c]{ReLU}}}
		\quad
		{\color{blue}\rotatebox{90}{\framebox[3cm][c]{FC : $4096 \times 500$}}}
		{\color{magenta}\rotatebox{90}{\framebox[3cm][c]{BatchNorm}}}
		{\color{cyan}\rotatebox{90}{\framebox[3cm][c]{ReLU}}}
		\quad
		{\color{blue}\rotatebox{90}{\framebox[3cm][c]{FC : $4096 \times 3$}}}
		{\color{cyan}\rotatebox{90}{\framebox[3cm][c]{$\pi \tanh$}}}
		\caption{Pose Network for the axis-angle representation}
		\label{fig:pose_network}
	\end{figure}

	The proposed network is a modification of the VGG-M network \cite{Chatfield:BMVC14} and has two parts, a feature network and a pose network, as illustrated in Fig.~\ref{fig:overall_network}.
	The feature network is identical to the VGG-M upto layer FC6 and is initialized using pre-trained weights, learned by \cite{Chatfield:BMVC14} for the ImageNet classification task \cite{ImageNet}. The pose network takes as input the output of the feature network and has 3 fully connected layers with associated activations and batch normalization as outlined in Fig. ~\ref{fig:pose_network}. The feature network is shared across all object categories but each category has its own pose network. Note that this is similar to \cite{Tulsiani:CVPR15, Su:ICCV15} except that we branch out at FC6 whereas they branch at FC7. Also note that, we take as input the class id of the image, which tells us the corresponding pose network to select for output pose.
	
	For the axis-angle representation, the output of the pose network is $\theta v$ and we 
	model the constraints $\theta \in [0, \pi)$ and $v_i \in [-1, 1]$ using a $\pi \tanh$ 
	non-linearity. An additional advantage of modeling pose in the continuous domain is that we can 
	now use the more appropriate geodesic loss instead of the cross entropy loss for 
	pose-classification or the mean squared error for standard regression. We optimize the geodesic 
	error between the ground-truth rotation $R$ and the estimated rotation $\hat{R}$, given by 
	$\mathcal{L} = d_A(R, \hat{R})$ from Eqn.~\eqref{eq:distR}. For the quaternion representation, the output of the network is now 4-dimensional and the unit-norm constraint is enforced by choosing the non-linearity as L2 normalization. The corresponding loss function $\mathcal{L} = d_Q(R, \hat{R})$ is obtained from Eqn.~\eqref{eq:distQ}. 
	
	\subsection{Data Augmentation by 3D Pose Jittering}
	\label{sec:augmentation}
\begin{figure}[h]
		\centering
		\begin{subfigure}{0.31\linewidth}
			\includegraphics[height=2cm, width=\linewidth]{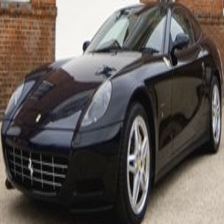}
			\caption{original}
		\end{subfigure}
		~
		\begin{subfigure}{0.31\linewidth}
			\includegraphics[height=2cm, width=\linewidth]{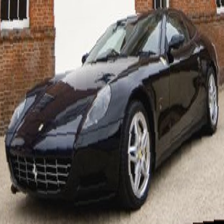}
			\caption{$\delta ct: +4^\circ$}
		\end{subfigure}
		~
		\begin{subfigure}{0.31\linewidth}
			\includegraphics[height=2cm, width=\linewidth]{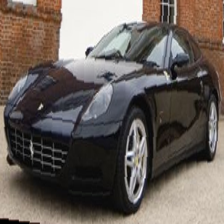}
			\caption{$\delta ct: -4^\circ$}
		\end{subfigure}
		
		\begin{subfigure}{0.31\linewidth}
			\includegraphics[height=2cm,width=\linewidth]{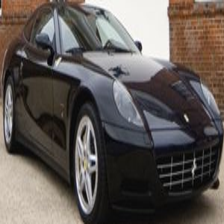}
			\caption{flipped}
		\end{subfigure}
		~
		\begin{subfigure}{0.31\linewidth}
			\includegraphics[height=2cm, width=\linewidth]{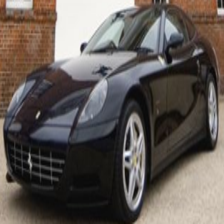}
			\caption{$\delta az: +2^\circ$}
		\end{subfigure}
		~
		\begin{subfigure}{0.31\linewidth}
			\includegraphics[height=2cm, width=\linewidth]{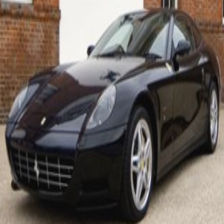}
			\caption{$\delta az: -2^\circ$}
		\end{subfigure}
		\caption{Augmented training samples from a car image}
		\label{fig:data_augmentation}
	\end{figure}

	We assume that each image is annotated with a 3D rotation $R(az, el, ct) = R_Z(ct) R_X(el) R_Z(az)$, where $R_Z$ and $R_X$ denote rotations around the $z$- and $x$-axis respectively. Jittered bounding boxes (bounding boxes with translational shifts that have sufficient overlap with the original box), like in V\&K \cite{Tulsiani:CVPR15}, introduce small unknown changes in the corresponding $R$. Instead, we augment our data by generating new samples corresponding to known small shifts in camera-tilt and azimuth. We call this new augmentation strategy \emph{3D pose jittering} (see Fig.~\ref{fig:data_augmentation}). Small shifts in camera-tilt lead to in-plane rotations, which are easily captured by rotating the image. Small shifts in azimuth lead to out-of-plane rotations, which are captured by homographies estimated from 2D projections of 3D point clouds corresponding to the object. We generate a dense grid of samples corresponding to $R(az \pm \delta az, el, ct \pm \delta ct)$. We also flip all samples, which corresponds to $R(-az, el, -ct)$. 

	Along with these augmented images, we also use rendered images provided publicly by Render-for-CNN \cite{Su:ICCV15} \footnote{https://shapenet.cs.stanford.edu/media/syn\_images\_cropped\_bkg\_overlaid.tar}  to supplement our training data. We present an analysis in \S\ref{sec:analysis} which shows that using these rendered images is important to reduce the problem of ``un-seen'' and ``under-seen views''.
	
	\begin{table*}[ht]
		\centering
		\setlength{\tabcolsep}{1.5mm}
		\begin{tabular}{|c|cccccccccccc|c|}
			\hline
			Expt. & aero & bike & boat & bottle & bus & car & chair & dtable & mbike & sofa	& train & tv & Mean \\
			\hline
			V\&K \cite{Tulsiani:CVPR15} & \textbf{13.80} & \textcolor{red}{17.70} & \textbf{21.30} & 12.90 & 5.80 & 9.10 & \textcolor{red}{14.80} & \textcolor{red}{15.20} & \textbf{14.70} & 13.70 & 8.70 & \textcolor{red}{15.40} & \textcolor{red}{13.59} \\ 
			Render \cite{Su:ICCV15} & 15.40 & \textbf{14.80} & \textcolor{red}{25.60} & 9.30 & \textbf{3.60} & \textbf{6.00} & \textbf{9.70} & \textbf{10.80} & \textcolor{red}{16.70} & \textbf{9.50} & \textbf{6.10} & \textbf{12.60} & \textbf{11.67} \\ 
			\hline
			Ours (axis-angle) & \textcolor{red}{13.97} & 21.07 & 35.52 & \textbf{8.99} & \textcolor{red}{4.08} & \textcolor{red}{7.56} & 21.18 & 17.74 & 17.87 & \textcolor{red}{12.70} & 8.22 & 15.68 & 15.38 \\ 
			Ours (quaternion) & 14.53 & 22.55 & 35.78 & \textcolor{red}{9.29} & 4.28 & 8.06 & 19.11 & 30.62 & 18.80 & 13.22 & \textcolor{red}{7.32} & 16.01 & 16.63 \\ 
			\hline
		\end{tabular}
		\caption{A comparison of our framework with two state-of-the-art methods for the axis-angle and quaternion representations. We report the median geodesic angle error (lower is better). Best result in bold and second best in red (best seen in color).}
		\label{table:main_result}
	\end{table*}

	\section{Results and Discussion}
	\label{sec:results}
	In this section, we first discuss the dataset we use (\S\ref{sec:dataset}) and how we train our network (\S\ref{sec:training}). In \S\ref{sec:experiments}, we present an experimental evaluation of our framework using image patches inside ground-truth bounding box annotations of un-occluded and un-truncated objects in an image (same protocol as V\&K \cite{Tulsiani:CVPR15} - table 1 and Render-for-CNN \cite{Su:ICCV15} - table 2). In \S\ref{sec:analysis}, we provide an analysis of various decision choices we make, like: (i) depth of feature network, (ii) choice of feature network, (iii) choice of optimization strategy, (iv) using rendered images for data augmentation, and (v) finetuning the network. Finally, in \S\ref{sec:detected} we report performance using detected bounding boxes returned by Faster R-CNN \cite{Ren:FasterRCNN} under various metrics.
	
	\subsection{Dataset}
	\label{sec:dataset}
	For our experiments, we use the Pascal 3D+ dataset (release 1.1) \cite{Xiang:WACV14}, which has 3D 	pose annotations for 12 common categories of interest: aeroplane (aero), bicycle (bike), boat, bottle, bus, car, chair, diningtable (dtable), motorbike (mbike), sofa, train, and tvmonitor (tv). The annotations are available for both VOC 2012 \cite{PASCAL} and ImageNet \cite{ImageNet} images. We use ImageNet data for training, Pascal-train images as validation data and evaluate our models on Pascal-val images. For every training image, we generate roughly 162 augmented samples with shifts in the camera-tilt (from $-4^\circ$ to $+4^\circ$ in steps of $1^\circ$: x9), shifts in azimuth (from $-2^\circ$ to $+2^\circ$ in steps of $0.5^\circ$: x9) and flips (x2).
	
	\subsection{Training the Network}
	\label{sec:training}
	We train our network in two steps: (i) we train the pose network for every object category (keeping the feature network fixed) using augmented ImageNet trainval images as training data and Pascal-train images as validation data, and (ii) use this as the initialization to fine-tune the overall network with all object categories in an end-to-end manner using Pascal-train and ImageNet-trainval images with only flipped augmentation as our training data.  While training the pose networks, we first minimize the mean squared error (MSE) for 10 epochs and then minimize the geodesic viewpoint error (GVE) for 10 epochs. Our loss is non-linear with many local minima and minimizing the MSE allows us to initialize the weights for the GVE minimization problem. We use the Adam optimizer with a learning rate schedule of $10^{-3}/(1 + epoch)$. Our code was written in Keras \cite{chollet2015keras} with TensorFlow \cite{tensorflow2015-whitepaper} backend.
	
	\subsection{Experimental Evaluation}
	\label{sec:experiments}
	As mentioned earlier, we use ground-truth bounding boxes of un-occluded and un-truncated objects in an image to evaluate our framework. As in V\&K, we compute the geodesic angle between the ground-truth rotation and estimated rotation $d(R_1, R_2) = \frac{\|\log(R_1 R_2^T) \|_F}{\sqrt{2}}$ and present the median angle error (in degrees). We report the mean across three trials of the experiment corresponding to training the network from three different random initializations.
	
	As can be seen in Table~\ref{table:main_result}, we show competitive performance compared to V\&K and Render-for-CNN, getting the lowest error for 1 category and second lowest error for 6 categories. We are able to do this in spite of solving a harder problem, of estimating 3D pose in the continuous domain.
		
	\subsection{Ablative Analysis}
	\label{sec:analysis}
	In this section, we present five experiments that provide insight into various design choices for our framework. Experiments (i)-(iv) report results after training only the pose networks (with the feature network fixed). Experiment (v) discusses the effects of finetuning the overall network.
	
\begin{figure}[h]
		\centering
		\includegraphics[width=\linewidth]{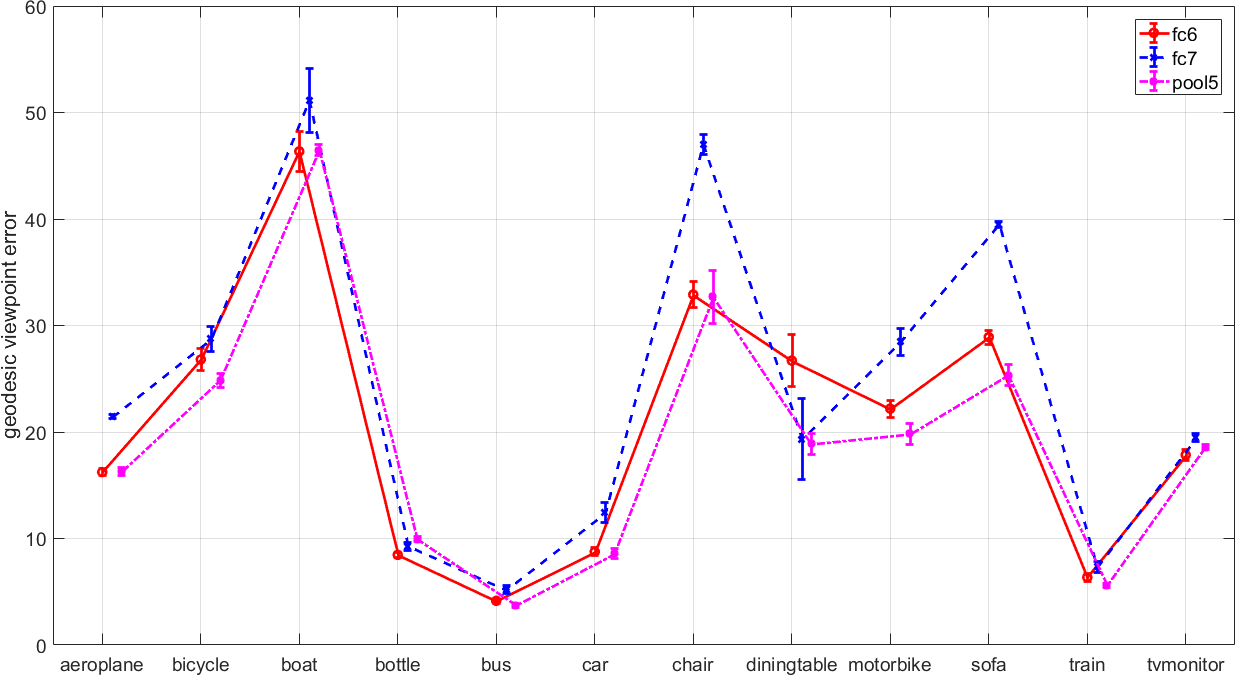}
		\caption{Median angle error for pose-networks trained with features extracted from the FC6, 
		FC7 and POOL5 layers of the VGG-M network (axis-angle representation)}
		\label{fig:fc6_fc7_pool5}
	\end{figure}

	\myparagraph{(i) Depth of Feature Network: FC6 vs FC7 vs POOL5}
	Our feature network is identical to the VGG-M network and uses the output at the FC6 layer as input to the pose networks. This is different from V\&K and Render-for-CNN which use the output at the FC7 layer of their feature networks. The rationale for using fewer layers is that a deeper network captures more invariances. This is because the VGG-M network is trained for classification of ImageNet Images, hence it is designed to be invariant to the object pose, which is a nuisance factor for the classification task. The question, however, is which layers of the network learn this invariance? The first few layers learn low-level features like edge detectors and simple shapes and deeper layers learn more complicated shapes. Similarly, we speculate that invariances like translation, color and scale are captured in the the convolutional layers, while pose invariances are learnt in the FC layers. Hence, features at FC7 are more invariant to pose compared to features at FC6 and POOL5. This is also borne out by the results in rows 5-7 of Table \ref{table:all_results} and Fig.~\ref{fig:fc6_fc7_pool5} where we see that the pose estimation error is less for networks trained with FC6 features compared to FC7 features for all categories except diningtable. This is consistent with the behaviour observed in \cite{Elhoseiny:ICML16} and \cite{Bakry:ICLR16}. Even though POOL5 results are slightly better than FC6, we branch at FC6 due to significant increase in computation for marginal increase in performance (POOL5 features are 18432 dimensional compared to 4096 dimensional FC6 features).
	
\begin{figure}[h]
		\centering
		\includegraphics[width=\linewidth]{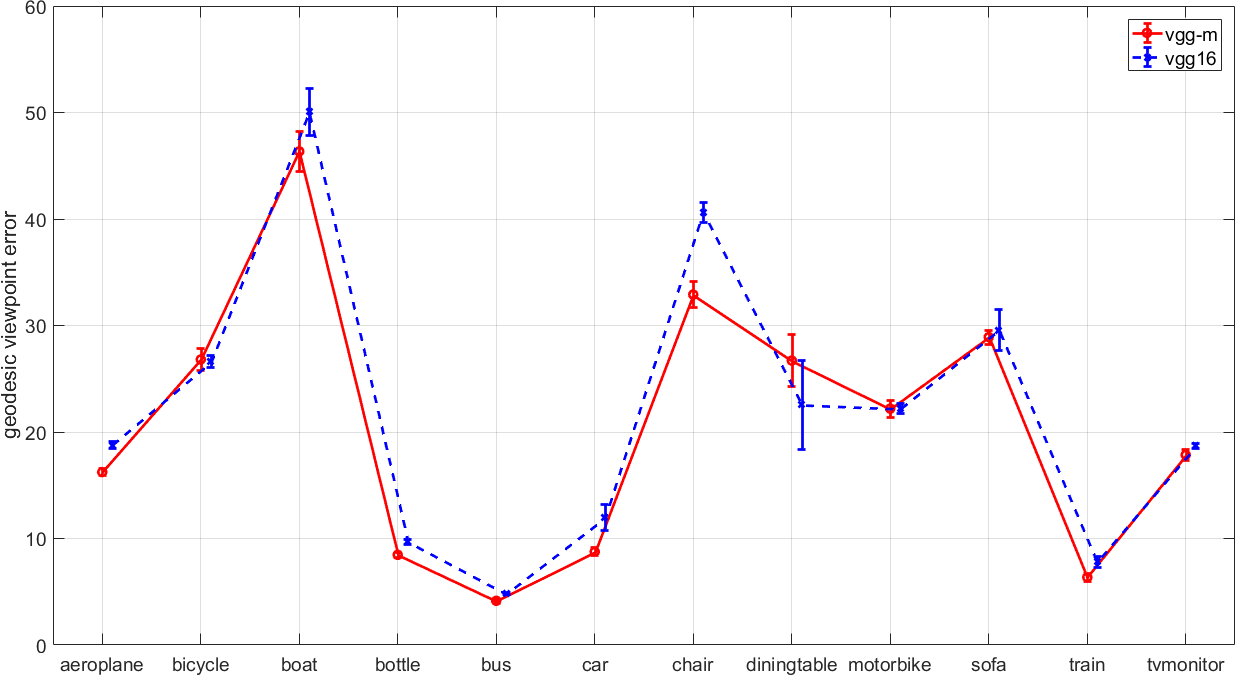}
		\caption{Median angle error under the VGG-M and VGG16 feature networks (axis-angle representation)}
		\label{fig:vggm_vgg16}
	\end{figure}

	\myparagraph{(ii) Type of Feature Network: VGG-M vs VGG16}
	Another decision choice is the use of the VGG-M network as our base-network. One could 
	exhaustively search over all possible choices of pre-trained networks to decide which network 
	is best suited for pose estimation. We chose not to do so, but compare the VGG-M and VGG16 
	networks which are two versions of the VGG architecture. We observe, in rows 8-9 of Table 
	\ref{table:all_results} and Fig.~\ref{fig:vggm_vgg16}, that the VGG-M network performs better than the VGG16 network. At the same time, we observe that pose estimation performance is not significantly affected by the choice of the feature network. Interestingly, augmenting training data with rendered images (explained later) worsens the performance of the VGG16 network (see rows 12 and 16 of Table~\ref{table:all_results}) whereas it improves the performance of the VGG-M network. 
	
\begin{figure}[h]
		\centering
		\includegraphics[width=\linewidth]{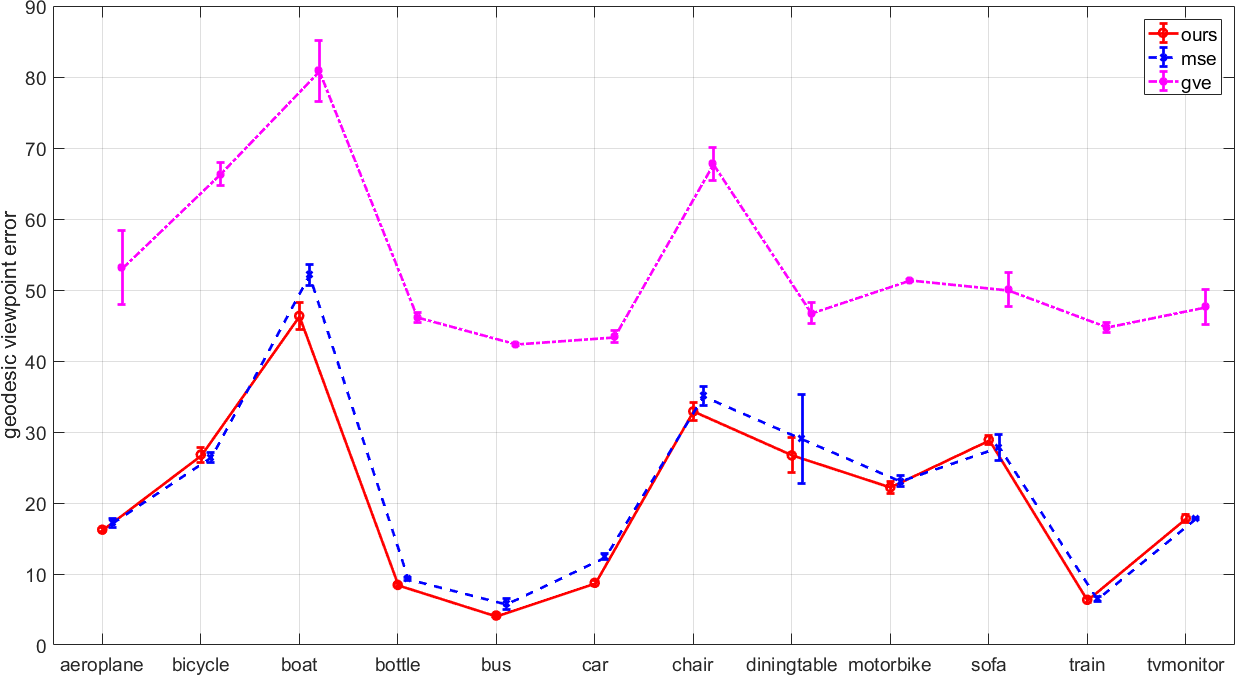}
		\caption{Median angle error under the different optimization strategies (axis-angle representation)}
		\label{fig:ours_mse_gve}
	\end{figure}

	\myparagraph{(iii) Optimization Strategy: MSE vs GVE vs Ours}
	As mentioned earlier, we minimize the MSE for 10 epochs and then minimize the GVE for 10 epochs. We do this to avoid the problem of local optima for the non-linear loss function and representation we use. We now show a comparison of what would happen if we just minimize the MSE for 20 epochs or the GVE for 20 epochs. As can be seen from Fig.~\ref{fig:ours_mse_gve} and Rows 8-11 of Table~\ref{table:all_results}, minimizing only the GVE leads us to bad local minima. However, initializing the GVE minimization with the result of the MSE minimization leads to significantly better performance. This phenomenon has also been observed in prior work on minimizing the geodesic distance in SO(3) \cite{Tron:CDC09}.

	\myparagraph{(iv) Data Augmentation using Rendered Images} 
	The number of images used for training, validation and testing are shown in Table~\ref{table:num_images}. For training, the number of augmented samples used is roughly $162$ times the number of images in the training set. We dig a little deeper into the viewpoint distribution to check if there are images in the training data that are `close' to the images present in the testing data. This is done using two metrics:
	\begin{align}
		\text{Cost1:} &\hspace{1em} \frac{1}{|\mathcal{D}_{Test}|} \sum_{i \in \mathcal{D}_{Test}} \min_{j \in \mathcal{D}_{Train}} d(\theta_i, \theta_j), \hspace{1em}\text{and} \\
		\text{Cost2:} &\hspace{1em} \frac{1}{|\mathcal{D}_{Test}|} \sum_{i \in \mathcal{D}_{Test}} \sum_{j \in \mathcal{D}_{Train}} \left [ d(\theta_i, \theta_j) < \epsilon \right ] .
	\end{align}
	Cost1 measures how close the nearest training sample is, in pose space, to each test sample. Cost2 on the other hand, measures how many training samples lie in an $\epsilon$ neighbourhood of each training sample. $\mathcal{D}_{Train}$ is the set of all original and flipped training images and $\mathcal{D}_{Test}$ is the set of all testing images. As can be seen by comparing Table \ref{table:viewpoint_dist} and row 3 of Table \ref{table:all_results}, we do well for categories that have many training examples in the $\epsilon$ neighbourhood of a test image, like bottle, bus, car and train, and don't do well for categories like bicycle, chair and motorbike that have few training examples in the $\epsilon$ neighbourhood. This is another way of saying that we do well for categories whose pose space is well sampled and worse for categories whose pose space is undersampled. Note that because we augment training images with small perturbations, the number of actual training samples close to a test sample will roughly be a multiple ($\sim 162$) of the entries in column 3 of Table~\ref{table:viewpoint_dist}.
	
	\begin{table}[h]
		\centering
		\begin{tabular}{|c|ccc|}
			\hline
			& \multicolumn{3}{|c|}{Pascal3D+} \\
			Category & Train & Val & Test \\
			\hline
			aeroplane & 1765 & 242 & 244 \\ 
			bicycle & 794 & 108 & 112 \\ 
			boat & 1979 & 177 & 163 \\ 
			bottle & 1303 & 201 & 177 \\ 
			bus & 1024 & 149 & 144 \\ 
			car & 5287 & 294 & 262 \\ 
			chair & 967 & 161 & 180 \\ 
			diningtable & 737 & 26 & 17 \\ 
			motorbike & 634 & 119 & 127 \\ 
			sofa & 601 & 38 & 37 \\ 
			train & 1016 & 100 & 105 \\ 
			tvmonitor & 1195 & 167 & 191 \\ 
			\hline
		\end{tabular}
		\caption{Number of images in Pascal3D+}
		\label{table:num_images}
	\end{table}
	
	\begin{table}[h]
		\centering
		\begin{tabular}{|c|@{\;}c@{\;}c@{\;}|@{\;}c@{\;}|}
			\hline
			Category & Cost1 & Cost2($\epsilon = 0.1$) & Cost2(+Rendered)\\
			\hline
			aeroplane & 0.047 & 23.12 & 1008.74\\ 
			bicycle & 0.051 & 11.18 & 950.622\\ 
			boat & 0.023 & 58.74 & 1801.09 \\ 
			bottle & 0.024 & 272.14 & 7733.42 \\ 
			bus & 0.011 & 168.19 & 6468.37 \\ 
			car & 0.012 & 217.21 & 3363.99 \\ 
			chair & 0.061 & 16.07 & 1124.23 \\ 
			diningtable & 0.026 & 39.71 & 2319.48 \\ 
			motorbike & 0.059 & 9.55 & 2319.48 \\ 
			sofa & 0.083 & 40.31 & 1733.97 \\ 
			train & 0.068 & 213.84 & 5639.89 \\ 
			tvmonitor & 0.029 & 74.15 & 3135.37 \\
			\hline
		\end{tabular}
		\caption{Viewpoint distribution under our two metrics}
		\label{table:viewpoint_dist}
	\end{table}

\begin{figure}[h]
		\centering
		\includegraphics[width=\linewidth]{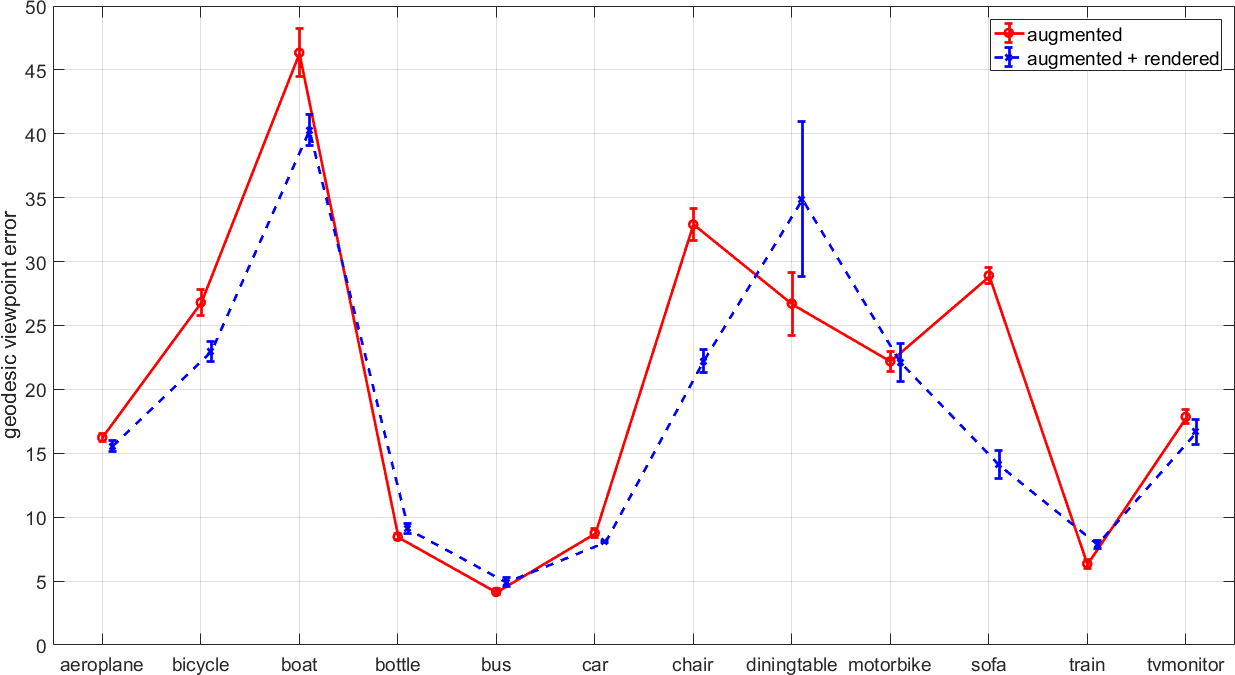}
		\caption{Median angle error under the axis-angle representation using rendered data}
		\label{fig:axis_angle_rendered}
	\end{figure}

\begin{figure}[h]
		\centering
		\includegraphics[width=\linewidth]{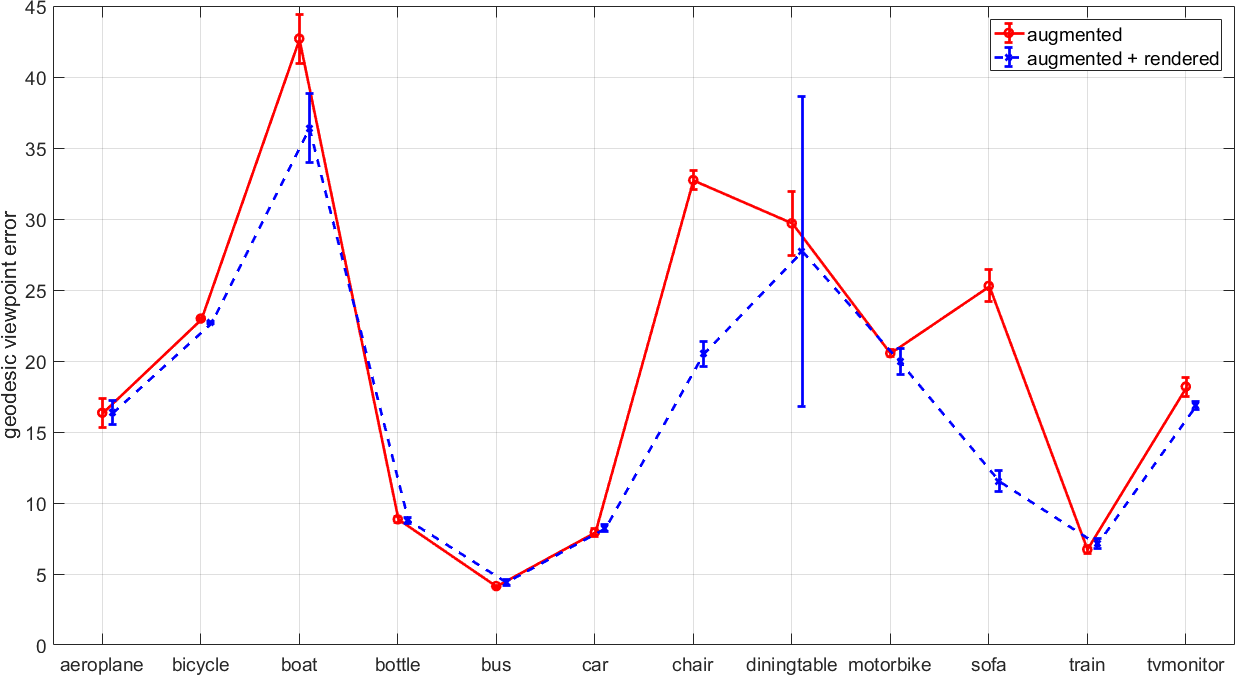}
		\caption{Median angle error under the quaternion representation using rendered data}
		\label{fig:quaternion_rendered}
	\end{figure}

	One way to increase the number of training examples and reduce this discrepancy of unseen poses is to use rendered images with known poses that sample the pose space in a more uniform manner. We use the rendered data made available by Render-for-CNN \cite{Su:ICCV15}. As can be seen in Figs.~\ref{fig:axis_angle_rendered} and \ref{fig:quaternion_rendered}, and rows 10-12 of Table \ref{table:all_results}, adding this rendered data helps reduce the errors significantly for categories like chair and sofa. This is observed for both the axis-angle and quaternion representations. Column 4 in Table \ref{table:viewpoint_dist} shows the updated Cost2 after including rendered images in $\mathcal{D}_{Train}$. Note that these numbers indicate neighbours in pose space and include images of varying sub-categories and appearances. Also, note that training purely on rendered images (row 14 of Table~\ref{table:all_results}) is worse than training on augmented data and training with both augmented and rendered data jointly gives best results.

\begin{figure}[h]
		\centering
		\includegraphics[width=\linewidth]{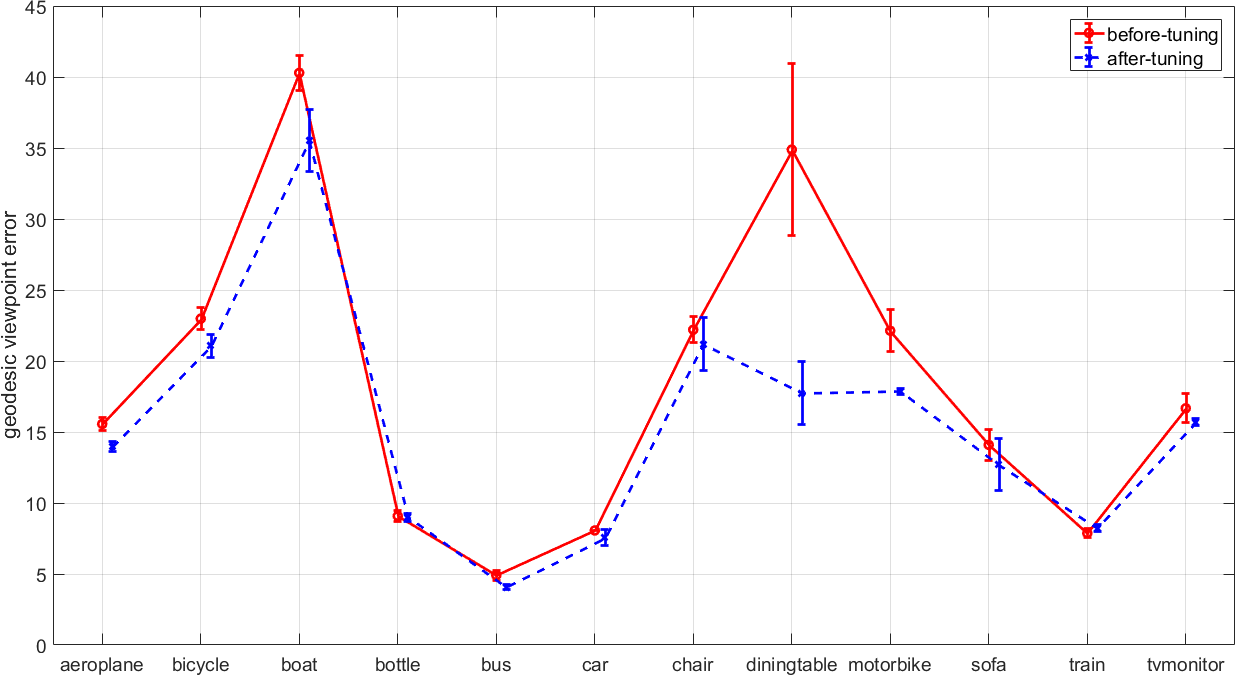}
		\caption{Median angle error under the axis-angle representation after fine-tuning the network}
		\label{fig:axis_angle_rendered_finetuned}
	\end{figure}

\begin{figure}[h]
		\centering
		\includegraphics[width=\linewidth]{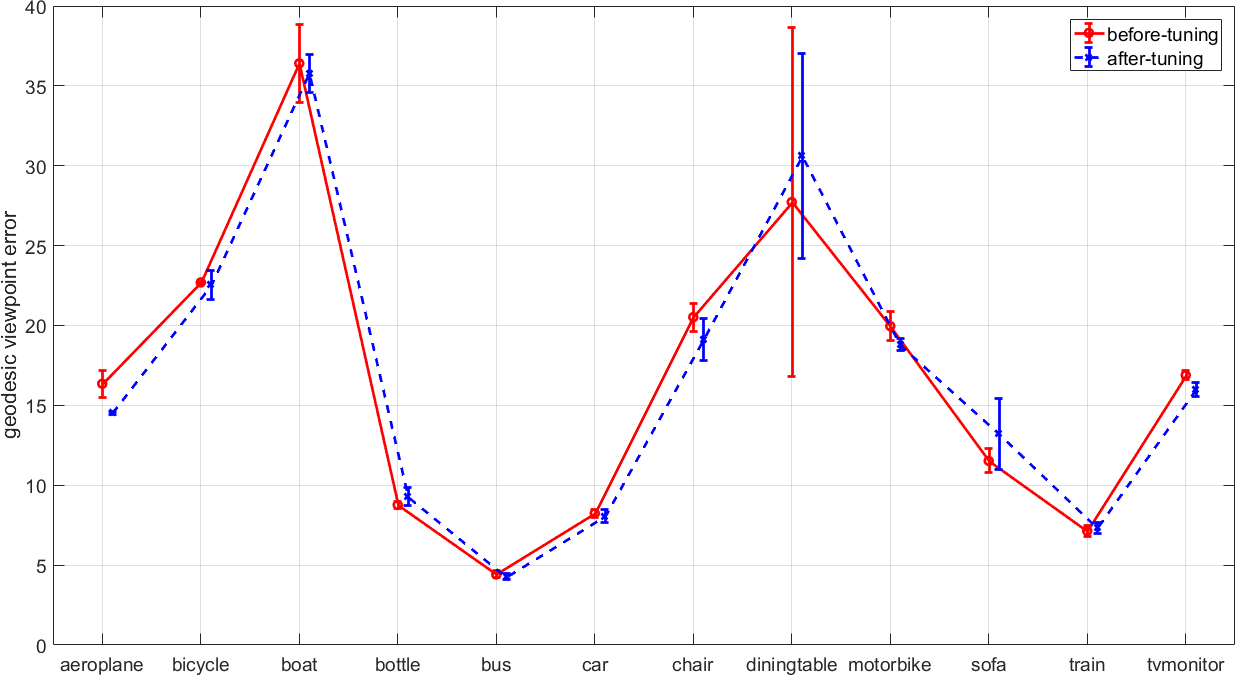}
		\caption{Median angle error under the quaternion representation after fine-tuning the network}
		\label{fig:quaternion_rendered_finetuned}
	\end{figure}

	\myparagraph{(v) Finetuning the Joint Network} 
	As mentioned earlier, we train our network in a two-step procedure. We first train all pose 
	networks with a fixed feature network and we then finetune the entire network. The finetuning  step updates the pre-trained feature network for the task of pose regression. We minimize the 
	geodesic viewpoint error for 30 epochs using the Adam optimizer with original and flipped 
	images of ImageNet trainval and Pascal train images. We use a weighted loss inversely proportional to the number of images per object category. For the axis-angle representation, we used a learning rate of $10^{-5}$ and find an improvement of $\sim 3^\circ$ in the median angle error averaged across all object categories, comparing rows 16 and 20 of Table \ref{table:all_results} and Fig~\ref{fig:axis_angle_rendered_finetuned}. For the quaternion representation, the optimization converges at a lower learning rate of $10^{-6}$, but doesn't	show a significant improvement after finetuning, comparing rows 17 and 21 of Table \ref{table:all_results} and Fig.~\ref{fig:quaternion_rendered_finetuned}.
	
	\subsection{Using Detected bounding boxes} 
	\label{sec:detected}
\begin{figure}
		\centering
		\includegraphics[width=\linewidth]{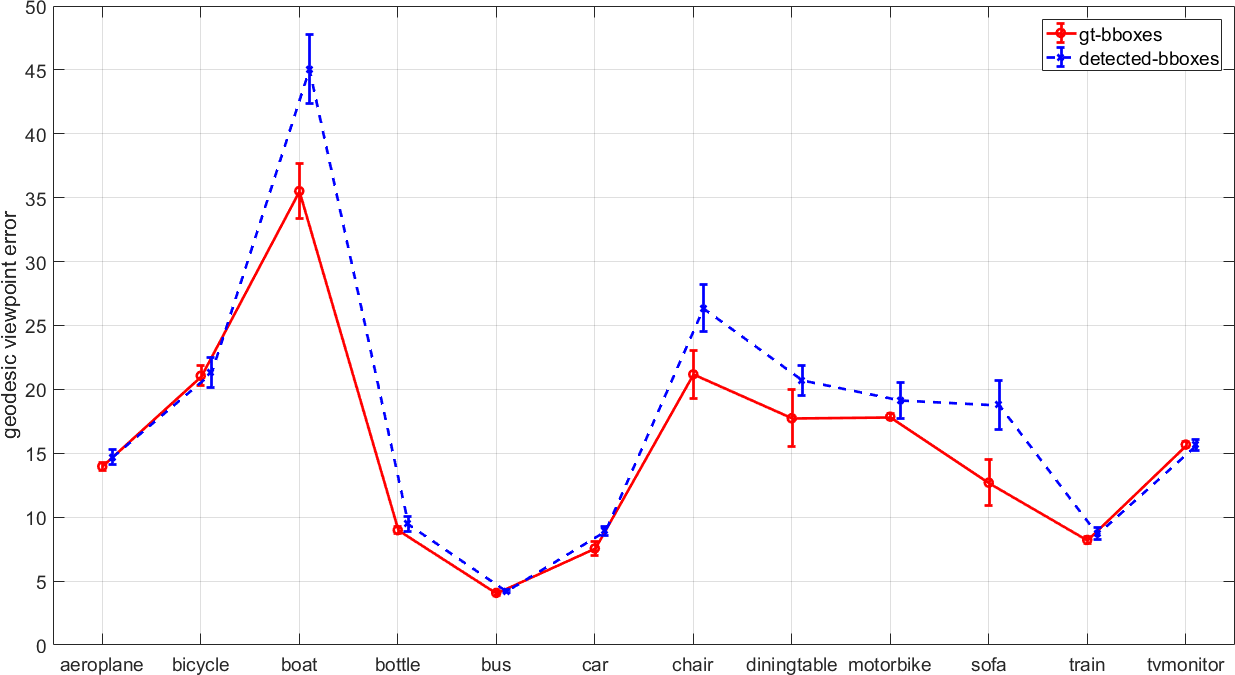}
		\caption{Median angle error under the axis-angle representation with ground-truth and detected bounding boxes}
		\label{fig:detected}
	\end{figure}

	The results presented so far have been obtained with ground-truth bounding boxes for un-occluded and un-truncated objects. We now present results on detected bounding boxes. We run the Faster R-CNN \cite{Ren:FasterRCNN} detection code to get bounding boxes for test images and then run our trained models on the patches extracted from these bounding boxes to get a corresponding pose. For every ground-truth bounding box (with annotated 3D pose), we find the detected box with the largest intersection-over-union overlap and compute the median angle error between the ground-truth pose and the estimated pose. As can be seen from Fig.~\ref{fig:detected} and Table \ref{table:det_results}, we lose performance slightly $\sim 2^\circ$ in going from ground-truth bounding boxes to detected bounding boxes. We also compare the performance of our method with V\&K \cite{Tulsiani:CVPR15} under the $ARP_{\theta}$ metric which requires sufficient overlap (intersection over union $> 0.5$) between detected and ground-truth bounding box as well closeness between ground-truth and predicted 3D pose, $\Delta(R_{gt}, R_{pred}) < \theta$. For this experiment, we use the detections provided publicly by V\&K \footnote{http://www.cs.berkeley.edu/~shubhtuls/cachedir/vpsKps/VOC2012\_val\_det.mat} for a direct comparison. We compare our performance with V\&K under the $ARP_{\pi/6}$ metric in Table~\ref{table:arp} and as can be seen, we perform slightly worse in all object categories.	We also compare under the AVP metric, which requires predicted azimuth $az$ to be close to the ground-truth azimuth, in Table~\ref{table:avp}. We perform slightly worse than Render-for-CNN and clearly worse than V\&K under this metric. However, we are at a disadvantage here, because the other methods train networks that return azimuth labels directly for this experiment, whereas we still predict a continuous 3D pose, recover the azimuth angle from predicted rotation matrix and then bin it to get predicted azimuth label. Effectively, we're solving a much harder problem but still get comparable results.

	\section{Conclusion}
	\label{sec:conclusion}
	We have proposed a regression framework to estimate 3D object pose given a 2D image. We use axis-angle and quaternions to represent the 3D pose output by the CNN and minimize a geodesic loss function during training. We show competitive performance with current state-of-the-art methods and provide an analysis of different parts of our framework. 
	
	\myparagraph{Acknowledgments} This research was supported by NSF grant 1527340. The research project was conducted using computational resources at the Maryland Advanced Research Computing Center (MARCC). This work also used the Extreme Science and Engineering Discovery Environment (XSEDE) \cite{XSEDE}, which is supported by National Science Foundation grant number OCI-1053575.  Specifically, it used the Bridges system \cite{Bridges}, which is supported by NSF award number ACI-1445606, at the Pittsburgh Supercomputing Center (PSC).

	\begin{table*}[h!]
	\small
	\setlength{\tabcolsep}{1.75mm}
	\centering
	\begin{tabular}{|c|@{\;}c@{\;}|cccccccccccc|c|}
		\hline
		\# & Expt. & aero & bike & boat & bottle & bus & car & chair & dtable & mbike & sofa & train & tv & Mean \\
		\hline
		
		1 & \cite{Tulsiani:CVPR15} & 13.80 & 17.70 & 21.30 & 12.90 & 5.80 & 9.10 & 14.80 & 15.20 & 14.70 & 13.70 & 8.70 & 15.40 & 13.59 \\ 
		2 & \cite{Su:ICCV15} & 15.40 & 14.80 & 25.60 & 9.30 & 3.60 & 6.00 & 9.70 & 10.80 & 16.70 & 9.50 & 6.10 & 12.60 & 11.67 \\ 
		\hline

		3 & axis-angle & 16.24 & 26.81 & 46.35 & 8.47 & 4.15 & 8.76 & 32.90 & 26.71 & 22.20 & 28.91 & 6.36 & 17.85 & 20.48 \\ 
		4 & quaternion & 16.35 & 22.99 & 42.71 & 8.85 & 4.15 & 7.93 & 32.74 & 29.70 & 20.55 & 25.29 & 6.73 & 18.20 & 19.68 \\ 
		\hline

		5 & fc6 & 16.24 & 26.81 & 46.35 & 8.47 & 4.15 & 8.76 & 32.90 & 26.71 & 22.20 & 28.91 & 
		6.36 & 17.85 & 20.48 \\ 
		6 & fc7 & 21.45 & 28.75 & 51.13 & 9.26 & 5.19 & 12.42 & 47.00 & 19.34 & 28.50 & 39.49 
		& 7.34 & 19.47 & 24.11 \\ 
		7 & pool5 & 16.30 & 24.85 & 46.46 & 9.93 & 3.72 & 8.56 & 32.68 & 18.91 & 19.81 & 25.33 & 5.59 & 18.57 & 19.23 \\ 
		\hline

		8 & mse(10)+gve(10) & 16.24 & 26.81 & 46.35 & 8.47 & 4.15 & 8.76 & 32.90 & 26.71 & 22.20 & 28.91 & 6.36 & 17.85 & 20.48 \\ 
		9 & mse(20) & 17.24 & 26.42 & 52.12 & 9.33 & 5.79 & 12.44 & 35.12 & 29.02 & 23.08 & 27.85 & 6.48 & 17.84 & 21.89 \\ 
		10 & gve(20) & 53.16 & 66.32 & 80.85 & 46.14 & 42.33 & 43.40 & 67.75 & 46.73 & 51.37 & 50.02 & 44.72 & 47.63 & 53.37 \\ 
		\hline
		
		11 & vggm & 16.24 & 26.81 & 46.35 & 8.47 & 4.15 & 8.76 & 32.90 & 26.71 & 22.20 & 
		28.91 & 6.36 & 17.85 & 20.48 \\ 
		12 & vgg16 & 18.76 & 26.62 & 50.07 & 9.69 & 4.81 & 11.97 & 40.62 & 22.55 & 22.20 & 
		29.56 & 7.81 & 18.71 & 21.95 \\ 
		\hline

		13 & augmented & 16.24 & 26.81 & 46.35 & 8.47 & 4.15 & 8.76 & 32.90 & 26.71 & 22.20 & 28.91 & 6.36 & 17.85 & 20.48 \\
		14 & rendered & 27.31 & 24.83 & 53.25 & 12.97 & 10.15 & 13.84 & 26.76 & 33.47 & 27.19 & 14.21 & 13.38 & 19.58 & 23.08 \\ 
		15 & both & 15.56 & 22.98 & 40.29 & 9.09 & 4.92 & 8.06 & 22.21 & 34.88 & 22.13 & 14.09 & 7.88 & 16.67 & 18.23 \\
		
		\hline
		16 & row3 + render & 15.56 & 22.98 & 40.29 & 9.09 & 4.92 & 8.06 & 22.21 & 34.88 & 22.13 
		& 14.09 & 7.88 & 16.67 & 18.23 \\ 
		17 & row4 + render & 16.35 & 22.70 & 36.41 & 8.77 & 4.42 & 8.24 & 20.53 & 27.73 & 19.96 & 11.53 & 7.14 & 16.89 & 16.72 \\ 
		18 & row6 + render & 19.43 & 29.76 & 49.25 & 9.37 & 5.85 & 10.89 & 35.14 & 30.06 & 26.69 
		& 20.06 & 8.82 & 17.44 & 21.90 \\ 
		19 & row12 + render & 19.65 & 27.61 & 49.26 & 9.85 & 4.89 & 12.13 & 46.66 & 30.76 & 
		23.12 & 36.80 & 8.71 & 19.72 & 24.10 \\ 
		\hline

		20 & row16 + finetune & 13.97 & 21.07 & 35.52 & 8.99 & 4.08 & 7.56 & 21.18 & 17.74 & 17.87 & 12.70 & 8.22 & 15.68 & 15.38 \\ 
		21 & row17 + finetune & 14.53 & 22.55 & 35.78 & 9.29 & 4.28 & 8.06 & 19.11 & 30.62 & 18.80 & 13.22 & 7.32 & 16.01 & 16.63 \\ 
		22 & row14 + finetune & 16.00 & 21.29 & 39.26 & 9.85 & 3.98 & 7.82 & 22.19 & 22.90 & 18.87 & 12.18 & 7.27 & 16.76 & 16.53 \\ 
		\hline
	\end{tabular}
	\caption{Median angle error under various experiments with ground-truth bounding boxes. Lower is better.}
	\label{table:all_results}
\end{table*}

\begin{table*}[h!]
	\small
	\centering
	\begin{tabular}{|c|cccccccccccc|c|}
	\hline
	Expt. & aero & bike & boat & bottle & bus & car & chair & dtable & mbike & sofa	& train 
	& tv & Mean \\
	\hline
	augmented & 18.59 & 26.43 & 56.47 & 9.13 & 4.31 & 10.05 & 41.83 & 27.00 & 22.19 & 27.60 & 
	7.06 & 19.23 & 22.49 \\ 
	+rendered & 17.38 & 23.32 & 54.11 & 10.10 & 5.22 & 9.39 & 25.45 & 21.98 & 20.88 & 18.13 & 
	8.27 & 17.78 & 19.33 \\ 
	+finetuned & 14.71 & 21.31 & 45.07 & 9.47 & 4.20 & 8.93 & 26.36 & 20.70 & 19.16 & 18.80 & 
	8.72 & 15.65 & 17.76 \\ 
	\hline
	\end{tabular}
\caption{Median angle error under various experiments with detected bounding boxes and 
axis-angle representation.}
\label{table:det_results}
\end{table*}

\begin{table*}[h!]
	\small
	\centering
	\begin{tabular}{|c|cccccccccccc|c|}
	\hline
	Expt & aero & bike & boat & bottle & bus & car &chair & dtable & mbike & sofa & train & tv & Mean \\
	\hline
	\cite{Tulsiani:CVPR15} & 64.0 & 53.2 & 21.0 & - & 69.3 & 55.1 & 24.6 & 16.9 & 54.0 & 42.5 & 59.4 & 51.2 & 46.5 \\
	Ours & 61.95 & 49.07 & 20.02 & 35.18 & 66.24 & 49.89 & 19.78 & 15.36 & 49.38 & 40.92 & 56.68 & 49.87 & 42.86 \\ 
	\hline
	\end{tabular}
	\caption{Comparision under the ARP metric for the results of the axis-angle + rendered + finetuned model. Higher is better.}
	\label{table:arp}
\end{table*}

\begin{table*}[h!]
	\small
	\centering
	\begin{tabular}{|c|cccccccccccc|c|}
		\hline
		Expt & aero & bike & boat & bottle & bus & car &chair & dtable & mbike & sofa & train & tv & Mean \\
		\hline
		\cite{Tulsiani:CVPR15}-4V & 63.1 & 59.4 & 23 & - & 69.8 & 55.2 & 25.1 & 24.3 & 61.1 & 43.8 & 59.4 & 55.4 & 49.1 \\
		\cite{Tulsiani:CVPR15}-8V & 57.5 & 54.8 & 18.9 & - & 59.4 & 51.5 & 24.7 & 20.4 & 59.5 & 43.7 & 53.3 & 45.6 & 44.5 \\
		\cite{Tulsiani:CVPR15}-16V & 46.6 & 42 & 12.7 & - & 64.6 & 42.8 & 20.8 & 18.5 & 38.8 & 33.5 & 42.4 & 32.9 & 36.0 \\
		\cite{Tulsiani:CVPR15}-24V & 37.0 & 33.4 & 10.0 & - & 54.1 & 40.0 & 17.5 & 19.9 & 34.3 & 28.9 & 43.9 & 22.7 & 31.1 \\
		\hline
		\cite{Su:ICCV15}-4V & 54.0 & 50.5 & 15.1 & - & 57.1 & 41.8 & 15.7 & 18.6 & 50.8 & 28.4 & 46.1 & 58.2 & 39.7 \\
		\cite{Su:ICCV15}-8V & 44.5 & 41.1 & 10.1 & - & 48.0 & 36.6 & 13.7 & 15.1 & 39.9 & 26.8 & 39.1 & 46.5 & 32.9 \\
		\cite{Su:ICCV15}-16V & 27.5 & 25.8 & 6.5 & - & 45.8 & 29.7 & 8.5 & 12.0 & 31.4 & 17.7 & 29.7 & 31.4 & 24.2 \\
		\cite{Su:ICCV15}-24V & 21.5 & 22.0 & 4.1 & - & 38.6 & 25.5 & 7.4 & 11.0 & 24.4 & 15.0 & 28.0 & 19.8 & 19.8 \\
		\hline
		Ours-4V & 52.43 & 50.80 & 19.74 & 35.66 & 61.24 & 46.82 & 20.85 & 20.31 & 50.60 & 42.01 & 53.42 & 53.11 & 42.25 \\ 
		Ours-8V & 42.98 & 37.96 & 13.18 & 34.61 & 41.59 & 38.66 & 16.13 & 12.55 & 37.94 & 33.19 & 43.00 & 40.43 & 32.68 \\ 
		Ours-16V & 29.90 & 24.37 & 7.73 & 32.06 & 38.75 & 29.23 & 12.18 & 10.32 & 25.62 & 24.82 & 29.50 & 25.16 & 24.14 \\ 
		Ours-24V & 21.71 & 14.21 & 5.62 & 29.44 & 29.16 & 25.15 & 9.16 & 6.98 & 18.94 & 15.47 & 26.38 & 17.97 & 18.35 \\ 
		\hline
	\end{tabular}
	\caption{Comparision under the AVP metric for the results of the axis-angle + rendered + finetuned model. Higher is better. 4/8/16/24V refers to number of azimuth bins.}
	\label{table:avp}
\end{table*}

	\bibliographystyle{ieee}
	\bibliography{biblio/recognition,biblio/vidal,biblio/computationalresources}

\end{document}